\documentclass[conference]{IEEEtran}
\IEEEoverridecommandlockouts
\usepackage{graphicx}

\usepackage{comment,cuted}
\usepackage{amssymb}
\usepackage{multirow}
\usepackage{subcaption}
\usepackage{adjustbox}
\usepackage{color}
\usepackage{amsmath}
\usepackage{cite}
\usepackage{amsmath,amssymb,amsfonts}
\usepackage{algorithmic}
\usepackage{graphicx}
\usepackage{textcomp}
\usepackage{xcolor}
\def\BibTeX{{\rm B\kern-.05em{\sc i\kern-.025em b}\kern-.08em
    T\kern-.1667em\lower.7ex\hbox{E}\kern-.125emX}}
\begin{document}

\title{Semi-supervised Deep Representation Learning \\for Multi-View Problems\\
}

\author{\IEEEauthorblockN{{Vahid Noroozi}\textsuperscript{1},
        {Sara Bahaadini}\textsuperscript{2}, 
		{Lei Zheng}\textsuperscript{1} ,
        {Sihong Xie}\textsuperscript{3},
        {Weixiang Shao}\textsuperscript {4} , 
        {Philip S. Yu}\textsuperscript{1}
        }
\IEEEauthorblockA{$^1$Department of Computer Science, University of Illinois at Chicago, IL, USA  \\
$^2$Department of Electrical Engineering and Computer Science, Northwestern University, IL, USA
\\
$^3$Department of Computer Science, Lehigh University, Pennsylvania, USA
\\
$^4$Google Inc., Mountain View, California, USA\\
email:vnoroo2@uic.edu, sara.bahaadini@u.northwestern.edu, lzheng21@uic.edu, sxie@cse.lehigh.edu, \\software.shao@gmail.com, psyu@uic.edu \\
}
}

\maketitle

\begin{abstract}
While neural networks for learning representation of multi-view data have been previously proposed as one of the state-of-the-art multi-view dimension reduction techniques, how to make the representation discriminative with only a small amount of labeled data is not well-studied. We introduce a semi-supervised neural network model, named Multi-view Discriminative Neural Network (MDNN), for multi-view problems. MDNN finds nonlinear view-specific mappings by projecting samples to a common feature space using multiple coupled deep networks. It is capable of leveraging both labeled and unlabeled data to project multi-view data so that samples from different classes are separated and those from the same class are clustered together. It also uses the inter-view correlation between views to exploit the available information in both the labeled and unlabeled data. Extensive experiments conducted on four datasets demonstrate the effectiveness of the proposed algorithm for multi-view semi-supervised learning.
\end{abstract}


\section{Introduction}
\label{intro}
In many real-world problems, more than one set of features, referred to as views of the data, are available. For example, a web page can be represented by text data, images, and meta-data. Multiple views can help improve the performance of many learning tasks because each view can provide information complementary to others, and learning using all views can maximally exploit the information available. In particular, multi-view dimension reduction has been proven effective for learning from high dimensional multi-view data~\cite{xu2013survey} such as image and text in image processing~\cite{Chaudhuri2009}, speech and video in speech processing \cite{Arora2013}, and multilingual texts in language processing~\cite{Faruqui2014}.

Compared to traditional multi-view dimension reduction, multi-view dimension reduction using deep networks has shown the state-of-the-art results~\cite{lecun2015deep}. Projections are learned to map all views to a common feature space where view information is retained and fused. The new space can enable or improve learning algorithms that are not applicable or inferior in multiple high-dimensional spaces.
Low-dimensional representations learned in such a way are without labeled data and thus not sufficiently discriminative for end tasks such as classification and clustering.
Discriminative multi-view dimension reductions based on CCA~\cite{wang2015deep}, topic models~\cite{Zhu2009}, and information bottleneck~\cite{Xu2014} can help learn representations that can not only unify different views by dimensionality reduction but also discriminate different classes. However, as the networks become deeper,
more parameters need to be learned and a larger amount of labeled data are required which is not readily available in many applications. The high cost of obtaining labeled data along with the growing size of unlabeled data has driven the development of semi-supervised learning that combines labeled and unlabeled data to mitigate the issue. However, there is still a lack of a semi-supervised deep discriminative method for multi-view dimension reduction.

\begin{figure*}[t!]
    \centering
    \includegraphics[width=0.78\textwidth]{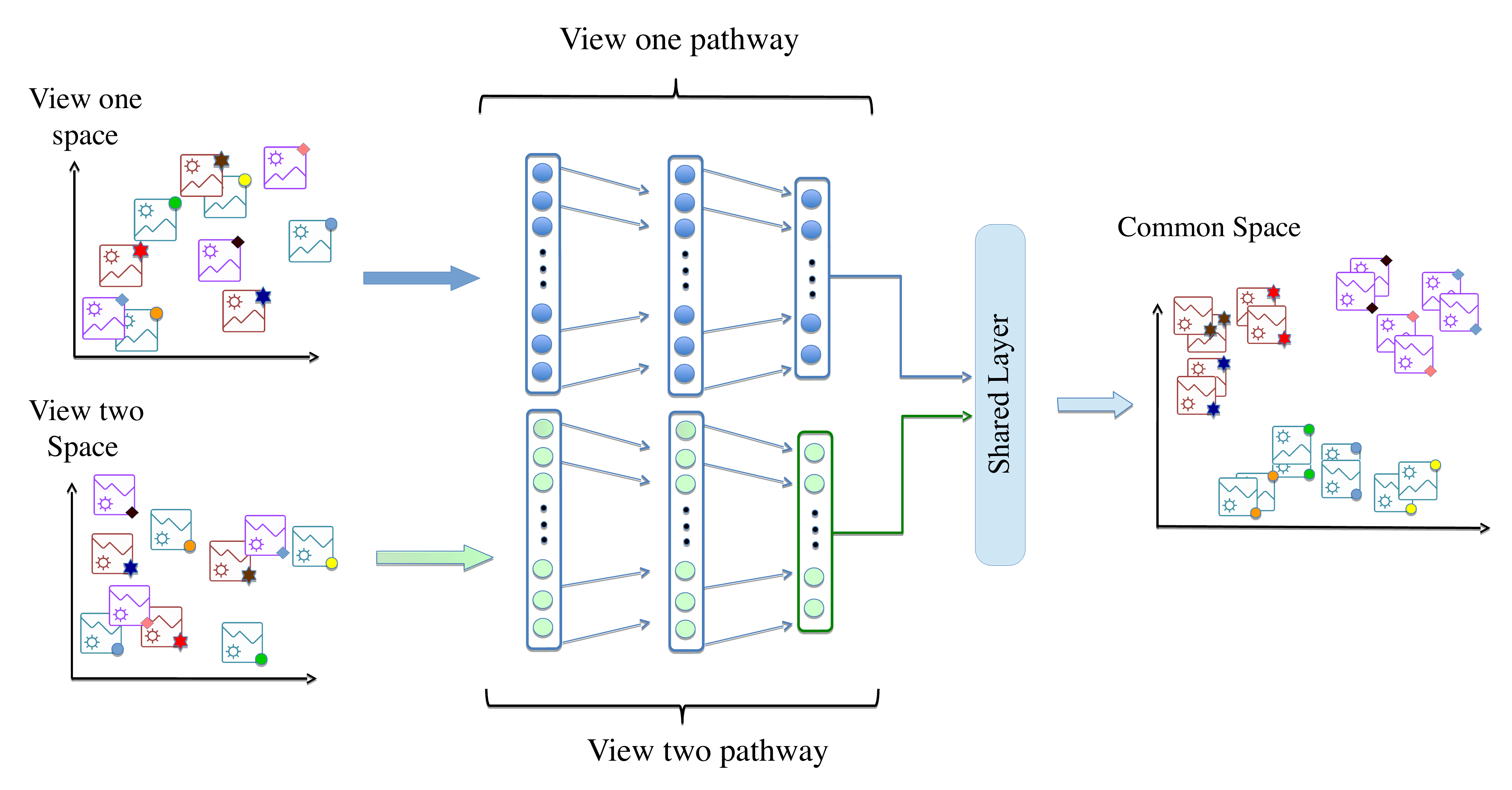}
\caption{MDNN structure for two views. In the left side of the picture, instances are shown in their original input space for both views. After passing them through MDNN, the common feature space is obtained. It is depicted on the right side of the picture. In this example, samples belong to three classes. The color indicates the class of an instance. Corresponding views are marked with small shapes at the corner of each instance.}
    \label{fig:networks1}
\end{figure*}

We propose MDNN (Multi-view Discriminative Neural Network) for the above purpose, using only a small amount of labeled data and a large amount of unlabeled data.
MDNN maximizes between-class separations and minimizes within-class variations while leveraging label information for discriminativeness. MDNN consists of a pair of parallel neural networks coupled by a shared layer on the top of the last layers (see Fig.~\ref{fig:networks1}). The model is trained in a joint manner to find view-specific nonlinear transformations. The learned transformations are further used to project samples to the common space. MDNN not only projects paired instances from different views to the same space (maximal correlation) but also projects instances from different classes far from each other (inter-class separation) while instances with the same class label are close to each other (intra-class variation). 

To the best of our knowledge, MDNN is the first deep semi-supervised representation learning method in multi-view problems, which has all of the following properties in a single unified model: (i) yielding a discriminative feature representation, (ii) using the complementary information of other views to exploit the information in unlabeled data, and (iii) achieving the above properties using a large amount of unlabeled data to help learning with only a small amount labeled data.
We evaluate MDNN on four multi-view datasets, namely Noisy MNIST, WebKB, FOX, and CNN, and compare it to the state-of-the-art baselines. The proposed model is evaluated in cross-view learning setting where we have two views during the training but just one of them is available for test. Experimental results demonstrate that the proposed algorithm outperforms all the baselines in terms of accuracy especially when a limited number of labeled samples are available.
The reminder of this paper is organized as follows. In Section \ref{sec:pw}, an overview of the related previous works is given. The proposed algorithm is discussed in detail in Section \ref{sec:alg} and experiments are presented in Section \ref{sec:exp}. In Section \ref{sec:conc}, we conclude the paper.

\section{Previous Works}
\label{sec:pw}
In Table \ref{tab:cmp}, capabilities of different multi-view dimension reduction models are compared. The proposed algorithm (MDNN) is the only one that enjoys all the capabilities. CCA is a well-known dimension reduction technique for data with two views~\cite{Hardoon2004,Dhillon2011,Chaudhuri2009,Foster2008,Arora2013}. It finds two linear transformations to project the views to a common feature space, so that correlation between views is maximized. However, CCA suffers from the lack of nonlinearity in its transformations to model nonlinear data. Kernel CCA (KCCA) extends CCA to find nonlinear projections \cite{Scholkopft1999} for both views. KCCA requires training data during testing and does not easily scale to large datasets.

More recently, several deep neural network (DNN)-based algorithms have been proposed for nonlinear feature representation learning on multi-view problems \cite{ngiam2011multimodal,Schmidhuber2014}. A deep model for CCA estimation, referred to as deep CCA (DCCA), has also been proposed~\cite{andrew2013deep,benton2017deep}. Like CCA, DCCA is a parametric approach and is scalable to large datasets, and like KCCA, it can model nonlinearity in the data.
Nonetheless, linear CCA, KCCA and deep CCA are unsupervised feature learning techniques. They cannot exploit labels available (if any) during feature representation learning. The learned low-dimensional representations thus lack class discriminativeness that is critical to the success of end tasks such as classification and clustering. While discriminative representation learning from one-view using deep network or topic model has also been explored in~\cite{Xu2014,Zhu2009}.
Learning a discriminative representation from multi-view data can require more labeled data as the number of views and network layers increase. While semi-supervised techniques for deep learning has been explored in~\cite{Zhang2014,Ororbia2015} to use a large amount of unlabeled data to mitigate the lack of labeled data, semi-supervised discriminative multi-view learning has not been studied and is the focus of this paper.

Maximizing between-class separations while minimizing within-class variations have been widely used in many learning algorithms, such as Fisher's Linear Discriminant Analysis (FLDA)~\cite{Sugiyama2006}. However, FLDA is a linear technique and although kernel-based versions of LDA have been proposed (KLDA)~\cite{Scholkopft1999}, they suffer from similar drawbacks of CCA and KCCA, such as scalability and fixed kernels. A version of LDA based on neural networks has been introduced \cite{dorfer2015deep}. However, all these studies work with only a single view and do not benefit from the noise-robustness of CCA-based techniques, which is the result of maximizing the correlation between views.

\begin{table}[t]
\caption{Comparison of different techniques with MDNN on various aspects. MDNN is the only one that has all properties in a single unified model.}
\label{tab:cmp}
\centering
\begin{adjustbox}{max width=0.5\textwidth}
\begin{tabular}{l|l|l|l|l}
\textbf{Method}  & \textbf{Nonlinearity} & \textbf{Scalability} & \textbf{Discriminativity} & \textbf{Multi-view} \\ \hline
CCA  &    &    &   & \checkmark 
\\ \hline
KCCA  & $\checkmark$ (limited) &  &  & $\checkmark$  \\ \hline 
DCCA  & $\checkmark$  & $\checkmark$  & & $\checkmark$   \\ \hline
LDA &  & $\checkmark$ & $\checkmark$ & \\ \hline
KLDA  & $\checkmark$ (limited) &  & $\checkmark$  &  \\ \hline 
Deep LDA  & $\checkmark$ & $\checkmark$ & $\checkmark$  &  \\ \hline 
MDNN & $\checkmark$ & $\checkmark$   & $\checkmark$& $\checkmark$    \\ 
\end{tabular}
\end{adjustbox}
\end{table}

\section{The Proposed Algorithm}
\label{sec:alg}
The schematic representation of the proposed model for two views is shown in Fig. \ref{fig:networks1}. The MDNN comprises of two deep neural networks (one for each view) coupled in a shared layer (interview-layer). More networks can get coupled to handle more views. Both networks are trained jointly to find view-specific nonlinear transformations to map the input views to a common feature space. The inter-view layer encourages inter-view correlation between views and is responsible for exploiting the information in both views of both labeled and unlabeled. All views of a single instance are projected as near as possible to each other. Moreover, two objectives are imposed on the output layer of each view independently to make the new space discriminative. It is achieved by maximizing intra-view discrimination using the labeled data: instances of the same class in one view are mapped closed together, whereas instances of different classes are mapped distant apart. Such properties make all the views of each instance to be \textit{highly correlated}, and instances of different classes are \textit{easily separable}.

These two parts of the model work in a joint manner to learn the desired representation from all the labeled and unlabeled data, and each can be considered as a regularizer for the other during the subspace learning. We train our model with backpropagation to learn two nonlinear transformations through optimizing the introduced objective functions. After training, the network is employed to map multiple views of data to a common low-dimensional space, where classifiers can be trained.

The purpose of using an independent network for each view is to learn low-level view-specific representations according to the properties of each view. Thus, the architecture of each network, such as the type or number of layers, can get adjusted according to the view's properties. In addition, representations obtained from higher levels of the networks are more likely to reveal the views' statistical properties compared to the original inputs \cite{srivastava2012learning}. 

\subsection{Deep Model Definition}
For a two-view problem, the training set is represented as ${\cal{X}} = \big\{(x_1^i,x_2^i)|x_1^i \in X_1, x_2^i \in X_2, 1<i<N\big\}$, where $(x_1^i,x_2^i)$ is a training sample with views $x_1^i \in \mathbb{R}^p$ and $x_2^i \in \mathbb{R}^q$ with dimension $p$ and $q$, respectively. $N=L+U$ is the total number of training pairs consisting of $L$ labeled and $U$ unlabeled pairs. The label set for labeled samples is denoted by ${\cal{Y}} = \{y^i|1 \le i \le L\}$. 

\begin{figure}[b]
    \centering
  \includegraphics[width=0.45\textwidth]{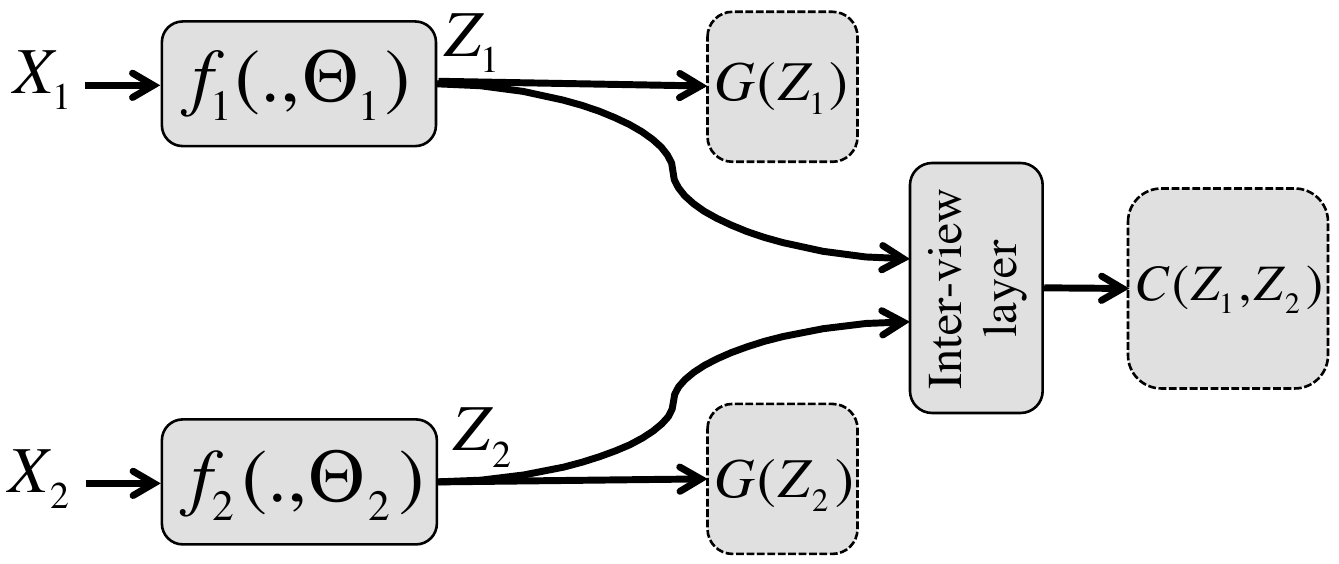}
\caption{A batch of instances are denoted for view one and two with $X_1$ and $X_2$, respectively. They are passed through non-linear view-specific functions $f_1$ and $f_2$. Discriminativity is imposed through the objective function $G(.)$ over the outputs $Z_1$ and $Z_2$. The maximization of inter-view correlation is imposed through the objective $C(Z_1,Z_2)$.}
\label{fig:networks2}
\end{figure}

We aim to learn two nonlinear view-specific functions ${f_1}(x;\Theta_1):{X_1} \to {Z_1}$ and ${f_2}{(x;\Theta_2)}:{X_2} \to {Z_2}$ that map the given paired views to the embedding spaces $Z_1$ and $Z_2$. Slightly abusing the notation, inputs to the first layers of the networks for the two views are batches of samples denoted by $X_1$ and $X_2$, and the hidden representations output by the last layers right before the shared layer are denoted by $Z_1$ and $Z_2$. Parameters $\Theta_1$ and $\Theta_2$ are the parameters of the two networks, respectively (see Fig. \ref{fig:networks2}).

\subsection{Objective Function}
\label{sec:obj}
To learn a discriminative representation for more effective classification, we define the objective function
\begin{equation}
\begin{aligned}
L({Z_1},{Z_2};{\theta _1},{\theta _2}) &= C({Z_1},{Z_2}) +&\\&{\lambda}(G({Z_1}) + G({Z_2}))- \alpha ({\left\| {{\theta _1}} \right\|^2} + {\left\| {{\theta _2}} \right\|^2})
\end{aligned}
\end{equation}
\noindent
where the function $C\left( {{Z_1},{Z_2}} \right)$ maximizes the inter-view correlation between the samples in the new space,
and functions $G(Z_i)$ encourages discriminative subspaces.
The term ${\left\| {{\Theta_1}} \right\|^2} + {\left\| {{\Theta _2}} \right\|^2}$ with regularization parameter $\alpha$ is added to regularize the networks. Parameter $\lambda$ specifies the trade-off between the importance of the inter-view correlation and intra-view discrimination properties in the new space.

We define the function $C$ based on CCA, which maps multiple views of samples into a new space where paired views of
each sample are highly correlated using a linear transformation matrix. It has been shown that the orthogonality of the learned dimensions is critical to effective representations of the multi-views~\cite{wang2015deep}. Considering the outputs of the two branches of MDNN as two sets of variables, also denoted by $Z_1$ and $Z_2$, CCA maximizes their correlation

\begin{equation}
\begin{aligned}
(v_1^*,v_2^*) &= \mathop {\arg \max }\limits_{{v_1},{v_2}} \textnormal{corr}(v_1^T{Z_1},v_2^T{Z_2}) &\\
&=\mathop {\arg \max }\limits_{{v_1},{v_2}} \frac{{v{{_1^T}_1}{\Sigma _{12}}{v_2}}}{{\sqrt {v_1^T{\Sigma _{11}}{v_1}v_2^T{\Sigma _{22}}{v_2}} }}
\end{aligned}
\end{equation}
\noindent
where $\Sigma _{ij}$ is the covariance matrix of $Z_i$ and $Z_j$:
\begin{equation}
{\Sigma _{ij}} = \frac{1}{{N - 1}}{\bar Z_i}\bar Z_j^T,
\end{equation}
where ${\bar Z_i}$ and ${\bar Z_j}$ are the centered matrices of $Z_i$ and $Z_j$, respectively.

Vectors $v_1^*$ and $v_2^*$ are the two linear transformation vectors that map $Z_1$ and $Z_2$ to a maximally correlated new space. Since such correlation function is invariant to scaling of transformation vectors $v_1$ and $v_2$, the objective function can be written as a constraint optimization problem as follows
\begin{equation}
(v_1^*,v_2^*) = \mathop {\arg \max }\limits_{v_1^T{\Sigma _{11}}{v_1} = v_2^T{\Sigma _{22}}{v_2} = 1} v_1^T{\Sigma _{12}}v_2^T
\end{equation}
We need to find other transformation vectors which produce projections uncorrelated with previous ones. The constrained problem to find all transformation vectors is
\begin{equation}
\mathop {(V_1^*,V_2^*) = }\limits_{} \mathop {\arg \max }\limits_{V_1^T{\Sigma _{11}}{V_1} = V_2^T{\Sigma _{22}}{V_2} = I} trace(V_1^T{\Sigma _{12}}{V_2})
\label{eq:cca_matrix}
\end{equation}
where matrices $V_1$ and $V_2$ contain transformation vectors as columns. Note that there are several ways to solve such optimization problems. It is shown in \cite{Hair2006} that the sum of the largest singular values of $R = \Sigma _{11}^{ - \frac{1}{2}}{\Sigma _{12}}\Sigma _{22}^{ - \frac{1}{2}}$ gives the maximal value of (\ref{eq:cca_matrix}),
and the corresponding eigenvectors are the optimal projection directions. The sum of all singular values can be estimated by the Frobenius matrix norm of $R$ as the following:

\begin{equation}
C({Z_1},{Z_2}) = {\left\| R \right\|_{F}} = \sqrt {trace({R^T}R)} 
\end{equation}
All covariance matrices in (\ref{eq:cca_matrix})
are regularized by a small positive number $r$ to ensure that the matrices are positive definite
\begin{equation}
{\Sigma _{ij}} = \frac{1}{{N - 1}}{\bar Z_i}\bar Z_j^T + rI.
\end{equation}

We define the function $G$ based on the two criteria of inter-class separation and intra-class variation to learn transformations that lead to a discriminative feature space. Inter-class separation measures how close instances from different classes are to each other. Intra-class variation measures how close instances from the same class are to each other. Generally, intra-class variation should be minimized while inter-class separation should be maximized to obtain a discriminative feature space.

The intra-class criterion $S_W$ (also referred to as a within-class scatter matrix) for a set of $L$ labelled samples from view $v$, $Z_v = \{z_v^1, z_v^2,..., z_v^L\}$, is defined as 
\begin{equation}
{S_W(Z_v)} = \frac{1}{L}\sum\limits_{i = 1}^{\left| C \right|} {{\sum\limits_{{z_v^j} \in {C_i}} {({z_v^j} - {m_v^i}){{({z_v^j} - {m_v^i})}^T}} }} 
\end{equation}
\noindent
where $|C|$ is the number of classes, and $z_v^j=f(d_v^j)$ is the $j_{th}$ instance of view $v$ in the new space. Variable $m_v^i$ denotes the mean of the samples from class $i$ for view $v$.

The inter-class criterion $S_B$ (also referred to as the between-class scatter matrix) for the same set of samples $Z_v$ can be defined as follows
\begin{equation}
{S_B(Z_v)} = \frac{1}{{2{L^2}}}\sum\limits_{i,j = 1}^{\left| C \right|} {{L_i}{L_j}({m_v^i} - {m_v^j}){{({m_v^i} - {m_v^j})}^T}}\nonumber
\end{equation}
\noindent
where $L_i$ is the number of labeled samples from class $i$. These two criteria can be merged into a single optimization problem as:
\begin{equation}
G(Z_v) = Tr\{ {({S_W}(Z_v) + {S_B}(Z_v) +rI)^{ - 1}}{S_B}(Z_v)\}\nonumber
\end{equation}
where $G(Z_v)$ measures the discriminiveness of the learned space for labeled samples of view $v$. The parameter $r$ is a regularization parameter to increase the stability of the inverse operation. Maximizing the function $G(Z_v)$ leads to maximizing $S_B(Z_v)$ and minimizing $S_W(Z_v)$ simultaneously to obtain a discriminative feature space.
\subsection{Optimization}
\label{subseq:opt}
To optimize the objective function $L$, we find the optimal values of all parameters for both networks, i.e., ${\Theta _1}$ and ${\Theta _2}$, using stochastic gradient descent(SGD). To use SGD, we split the samples into some labeled and unlabeled mini-batches. In labeled batches, labeled samples from each class are present proportional to their ratio in the whole data.

We estimate the gradient of $L$ with respect to the outputs of networks $Z_1$ and $Z_2$ to use the backpropagation technique. The backpropagation algorithm estimates other gradients to update the networks' parameters $\Theta_1$ and $\Theta_2$. 

If the singular value decomposition of matrix $R$ in function $C$ is $R = UDV$, then the gradient of function $C$ with respect to $Z_1$ can be estimated as follows:
\begin{equation}
\label{eq:z1der}
\frac{{\partial C({Z_1},{Z_2})}}{{\partial {Z_1}}} = \frac{1}{{N - 1}}(2{\nabla _{11}}{\bar Z_1} + {\nabla _{12}}{\bar Z_2})
\end{equation}
\noindent
where
\begin{equation}
\begin{aligned}
{\nabla _{12}} &= \Sigma _{11}^{ - \frac{1}{2}}U{V^T}\Sigma _{22}^{ - \frac{1}{2}}\nonumber&\\
{\nabla _{11}} &= \frac{{ - 1}}{2}\Sigma _{11}^{ - \frac{1}{2}}U{U^T}\Sigma _{11}^{ - \frac{1}{2}}
\end{aligned}
\end{equation}
$N$ denotes the total number of samples in the batch. Similar expressions hold for the gradient with respect to $Z_2$. More detail on calculating this gradient can be found in \cite{andrew2013deep}.

Calculating the gradient of the $G(Z_i)$ is not trivial. Similar variants of $G(Z_i)$ have been already investigated in other papers \cite{wu2017deep,dorfer2015deep}. In most cases, they tackled this optimization problem by reformulating it as a general eigen decomposition problem. We avoided such reformulation as we found out in our experiments that it increases the training instability of the neural networks. Therefore, we optimize $G(Z_i)$ without any reformulation by following \cite{stuhlsatz2012feature}.

We denote $S_W(Z_v)$ and $S_B(Z_v)$ as $S_W^v$ and $S_B^v$ respectively in the following derivations. If sample $z_v^n \in C_k$, then gradient of scatter matrices $S_W^v$ and $S_B^v$ for view $v$ are defined as \eqref{eq:sw} and \eqref{eq:sb}.

\begin{figure*}
\begin{equation}
\label{eq:sw}
\frac{{\partial S_W^v[i,j]}}{{\partial {Z_v}[n,p]}} = \frac{1}{L}\left\{ \begin{array}{l}
0\\
{Z_v}[j,p] - {m_v^k}[j]\\
{Z_v}[i,p] - {m_v^k}[i]\\
2({Z_v}[n,p] - {m_v^k}[n])
\end{array} \right.\begin{array}{*{20}{c}}
{{\rm{ if \quad i }} \ne n{\rm{ \quad and \quad j}} \ne n}\\
{{\rm{  if \quad i  =  n \quad and \quad j}} \ne n}\\
{{\rm{  if \quad i}} \ne n{\rm{ \quad and\quad j  =  n}}}\\
{{\rm{  if\quad i  =  n \quad and \quad j  =  n}}}
\end{array} 
\end{equation}

\begin{equation}
\label{eq:sb}
\frac{{\partial S_B^v[i,j]}}{{\partial {Z_v}[n,p]}} = \frac{1}{{{L^2}}}\sum\limits_{s = 1}^{\left| C \right|} {{L_s}} \left\{ \begin{array}{l}
0\\
{m_k}[j] - {m_v^s}[j]\\
{m_k}[i] - {m_v^s}[i]\\
2({m_k}[n] - {m_v^s}[n])
\end{array} \right.\begin{array}{*{20}{c}}
{{\rm{  if \quad i}} \ne n \quad{\rm{ and \quad j}} \ne {\rm{n}}}\\
{{\rm{  if \quad i  =  n\quad and \quad j}} \ne {\rm{n}}}\\
{{\rm{  if \quad i}} \ne n{\rm{ \quad and \quad j  =  n}}}\\
{{\rm{  if \quad i  =  n \quad and \quad j  =  n}}}
\end{array}
\end{equation}

\end{figure*}
\vspace{+1.3mm}
Defining $S_T=S_B + S_W$, then the gradient of the discriminative objective function $G(Z_v)$ is estimated as
{\small
\begin{align}
\frac{{\partial G({Z_v})}}{{\partial {Z_v}[n,p]}} &= Tr\{ {(S_T^v)^{ - 1}}S_B^v\} &\\ \nonumber&=\sum\limits_{s = 1}^d {\frac{{\partial {{(S_T^v)}^{ - 1}}[s,s]}}{{\partial {Z_v}[n,p]}}S_B^v[s,s]}  + {(S_T^v)^{ - 1}}[s,s]\frac{{\partial S_B^v[s,s]}}{{\partial {Z_v}[n,p]}}
\label{tag}
\end{align}}

As we can have the following \cite{petersen2008matrix}

\begin{equation}
\frac{{\partial {{(S_T^v)}^{ - 1}}S_B^v}}{{\partial {Z_v}}} =  - {(S_T^v)^{ - 1}}\frac{{\partial S_B^v}}{{\partial {Z_v}}}{(S_T^v)^{ - 1}}
\end{equation}

We rewrite the gradient $\frac{{\partial G(Z_v)}}{{\partial Z_v}}$ by using (\ref{eq:sw}) and (\ref{eq:sb}) as
\begin{equation}
\begin{aligned}
\frac{{\partial G(Z_v)}}{{\partial Z_v}} =& \frac{2}{{{L^2}}}[S_T^{ - 1}(Z_v){S_B}(Z_v) - I]S_T^{ - 1}(Z_v)(\sum\limits_{j = 1}^{\left| C \right|} {{M_v^j}} )\\
 -& \frac{2}{L}S_T^{ - 1}(Z_v){S_B}(Z_v)S_T^{-1}(Z_v)(Z_v - M_v)
\end{aligned}
\end{equation}
where
\begin{equation}
\begin{aligned}
M_v =& {({m_v^q}{.1^T})_{1 \le q \le \left| C \right|}}
\\
{M_v^j} =& {({L_j}.({m_v^j} - {m_v^q}){.1^T})_{1 \le q \le \left| C \right|}}
\end{aligned}
\end{equation}
\noindent
More detail can be found in \cite{stuhlsatz2012feature}. The gradient of the total objective function is used to train both the networks simultaneously with the backpropagation algorithm. It is necessary to train the model with mini-batches because the objective function is defined on the properties of whole space, not just a single instance. Therefore at each step, we need a batch of sample to optimize the objective function.

\section{Experiments}
\label{sec:exp}
In this section, we present the experimental evaluation and analysis of MDNN. All experiments are performed in cross-view classification setting, however, it can be extended to other tasks such as cross-modal image and text retrieval \cite{WangY0W016}. 
\subsection{Datasets}
\label{subsec:Datasets}
We evaluate the proposed algorithm on the following four datasets. A summary of the datasets is presented in Table \ref{tab:dataset}.

\begin{figure}[h]
\centering
\includegraphics[width=0.22\textwidth]{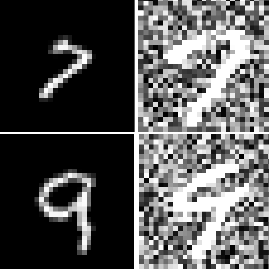}
\caption{Two examples from the multi-view Noisy MNIST. Left images are from view one, and right images are their corresponding samples from the second view.}
\label{fig:mnist}
\end{figure}

\textbf{Noisy MNIST}: Noisy MNIST is a noisy version of the well-known MNIST dataset that contains images of handwritten digits. Following \cite{wang2015deep}, a two-view version of MNIST for evaluating multi-view problem has been created. This was accomplished by rotating and adding random noise to the images of the dataset. Each image was rotated by a randomly sampled angle from a uniform distribution between $-pi/2$ and $+pi/2$. The resulting images were used as the first view. For each image, another image from the same class was selected randomly as the second view. Uniform noise samples in the range $[0, 1]$ were also added to each pixel of the images in the second view. The noisy MNIST dataset contains 70K grayscale images of digits 0 to 9. The split of 60,000/10,000 is used in the experiments for train/test. Two examples of this dataset are shown in Fig.~\ref{fig:mnist}.

\vspace{+1mm}
\textbf{Web Knowledge Base (WebKB)\footnote{http://vikas.sindhwani.org/manifoldregularization.html}}: It is a collection of 1,051 web documents crawled from four universities \cite{sindhwani2005beyond}. The data has two classes: course or non-course web pages. Each document has two views: 1) the textual content of the web page and 2) the anchor text on the links pointing to the web page.

\vspace{+1mm}
\textbf{CNN and FOX}: These two datasets were crawled from CNN and FOX web news \cite{qian2014unsupervised}. The category information extracted from their RSS feeds are considered as their class label. Each instance is represented in two views: the text view and image view. Titles, abstracts, and text body contents are considered as the text view data (view 1), and the image associated with the article is the image view (view 2). All text is stemmed by Porter stemmer, and l2-normalized TF-IDF is used as text features. Processed data samples in CNN and FOX datasets have 1,143 and 7,980 features respectively. Also, seven groups of color features and five textural features are used for image features \cite{qian2014unsupervised}, which results in 996 features for both datasets.

\begin{table}[t!]
\centering
\caption{Summary of the datasets.}
\label{tab:dataset}
\begin{tabular}{l|c|c|c}
Dataset & \# Instance & \# Feature & \# Class \\ \hline
Noisy MNIST & $70000$ & $784+784$ & $10$ \\
WebKB & $1051$ & $3000+1840$ & $2$ \\
FOX & $1523$ & $5477+996$ & $4$ \\
CNN & $2707$ & $7989+996$ & $7$
\end{tabular}
\end{table}

\begin{table*}[h!]
\centering
\caption{Performance of different methods trained with various number of labeled examples on Noisy MNIST and WebKB in terms of accuracy.}
\label{tab:mainres1}
\begin{tabular}{l|llll|llll}
\multicolumn{1}{c|}{Dataset} & \multicolumn{4}{c|}{Noisy MNIST} & \multicolumn{4}{c}{WebKB} \\
\# of labeled samples & $200$ & $400$ & $600$ & $All$ & $50$ & $100$ & $150$ & $All$  \\ \hline
MDNN  & $75.00$ & ${79.64}$ & ${80.32}$ & $\textbf{97.34}$ & $\textbf{86.58}$ & $\textbf{93.29}$ & $\textbf{94.46}$ & $\textbf{96.50}$  \\
Deep CCA & $80.17$ & $85.85$ & $77.91$ & $84.83$ & $80.17$ & $80.46$ & $93.00$ & $81.92$ \\
Deep LDA & $61.42$ & $73.29$ & $78.61$ & $96.83$ & $83.38$ & $86.88$ & $93.00$ & $95.62$  \\
Linear CCA  & $69.02$ & $72.70$ & $73.99$ & $76.13$ & $70.17$ & $71.62$ & $77.35$ & $83.50$ \\
 LDA & $40.05$ & $40.27$ & $25.65$ & $76.79$ & $57.43$ & $61.51$ & $62.23$ & $94.75$ \\
Kernel CCA & $\textbf{75.81}$ & $\textbf{87.25}$ & $\textbf{93.78}$ & $94.21$ & $77.46$ & $79.59$ & $80.17$ & $82.79$ 
\end{tabular}
\end{table*}

\begin{table*}[h!]
\centering
\caption{Performance of different methods trained with various number of labeled examples on FOX and CNN in terms of accuracy.}
\label{tab:mainres2}
{
\begin{tabular}{l|llll|llll}
\multicolumn{1}{c|}{Dataset} & \multicolumn{4}{c|}{FOX} & \multicolumn{4}{c}{CNN} \\
\# of labeled samples & $128$ & $256$ & $384$ & $All$ & $250$ & $500$ & $750$ & $All$  \\ \hline
MDNN  & $\textbf{73.62}$ & $\textbf{84.05}$ & $\textbf{85.43}$ & $92.51$ & $\textbf{78.59}$ & $\textbf{76.28}$ & $\textbf{80.08}$ & $\textbf{79.67}$  \\
Deep CCA & $59.44$ & $78.70$ & $75.19$ & $88.58$ & $44.85$ & $45.52$ & $45.39$ & $45.39$ \\
Deep LDA & $74.42$ & $80.07$ & $84.73$ & $\textbf{93.07}$ & $50.57$ & $63.49$ & $71.54$ & $70.48$  \\
Linear CCA  & $71.75$ & $77.16$ & $78.34$ & $82.57$ & $37.02$ & $41.23$ & $40.18$ & $45.45$\\
LDA & $70.66$ & $78.74$ & $81.69$ & $87.00$ & $65.71$ & $69.24$ & $73.98$ & $45.39$  \\
Kernel CCA & $71.25$ & $72.63$ & $71.06$ & $78.13$ & $38.20$ & $39.92$ & $40.84$ & $49.84$ 
\end{tabular}}
\end{table*}

\subsection{Baselines}
We compare the performance of MDNN with some the state-of-the-art algorithms for multi-view representation learning. From methods that do not use deep neural network, we compare MDNN to linear Canonical Correlation Analysis (CCA) and Kernel CCA (KCCA) as the most commonly used techniques for representation learning in multi-view problems \cite{xu2013survey,wang2015deep}. Although CCA finds linear transformations, it is still widely used because of its speed and simplicity. As traditional kernel CCA is not scalable, we use FKCCA \cite{lopez2014randomized} method which is an approximation of the real KCCA definition. 

Among all the DNN based approaches, deep CCA (DCCA) \cite{andrew2013deep} is used in the experiments because of the better performance than other DNN-based algorithms \cite{wang2015deep}. None of these CCA based techniques use the label information, and all are categorized as unsupervised feature reduction techniques.

Also, two approaches which consider label information are also selected as baselines, Linear Discriminant Analysis (LDA)~\cite{izenman2013linear}
and its neural network variant: Deep LDA \cite{dorfer2015deep}. These approaches are not designed for multi-view problems, but they are selected because they use labeled data to learn the new representation. Therefore they are applied on just the primary view, and cannot use inter-view relation between views.

\subsection{Experimental Settings}
\label{sebsec:es}
We perform all the experiments in cross-view learning, that is used in \cite{xu2013survey,wang2015deep,karen2016} for evaluating representation learning techniques in multi-view problems. In this setting, all views are available during the representation learning but one (view 2) is missing during the testing process. All the methods in the experiments use both primary and complementary views in the training process to learn a common feature space. After learning representation, primary view is mapped to the new learned space. Then a linear Support Vector Machine (SVM) classifier~\cite{svm} is trained on the new representation to evaluate it in a classification task. We would like to emphasize that the aim of this paper is to present a new representation for multi-view setting. Therefore, we selected linear SVM as the classifier instead of a more complicated method for classification. In this way, we can evaluate the effectiveness of the representation learning more accurately.

All the parameters are selected to obtain the best performance in cross-validation process. All neural network based models are trained for $150$ epochs. All samples are distributed randomly over the batches proportionally to their class size. 

Regularization parameter $\alpha$ is selected from $\{\textrm{1e-1, 1e-2, 1e-3, 1e-4, 1e-5}\}$ for all datasets. Trade-off parameter $\lambda$ is selected from $\{\textrm{1e-1, 1, 1e1, 1e2, 1e3, 1e4}\}$ for each dataset separately. Regularization $r$ is set to $1e-4$. Representation sizes are selected as the number of classes for each dataset except for WebKB which is set to $10$. These sizes may not give the best performance possible for MDNN, but they are set as the number of classes for all models to have a fair comparison among all techniques. Parameter $C$ of the SVM is also selected from $\{10, 1.0, 1e-1, 1e-2, 1e-3\}$ by cross-validation.

Networks with the same architecture consisting of $3$ hidden layers with the same number of hidden nodes are used for both views. The only exception is the network for WebKB which has 2 hidden layers instead of 3. Relu activation is used for all layers except the last one which has linear activation. Number of hidden nodes is selected as $1024$, $128$, $512$, $512$ for noisy MNIST, WebKB, FOX, and CNN datasets, respectively. We use a variant of SGD, called Adam \cite{kingma2014adam}, to optimize the neural networks. All the parameters of Adam are set as the its paper recommends. The architecture of the networks for all the neural network based models including deep CCA, Deep LDA, and MDNN are defined the same to have fair comparisons.

\begin{figure*}[h!]
\begin{center}
\begin{subfigure}{.38\textwidth}
  \centering
\includegraphics[width=.85\linewidth]{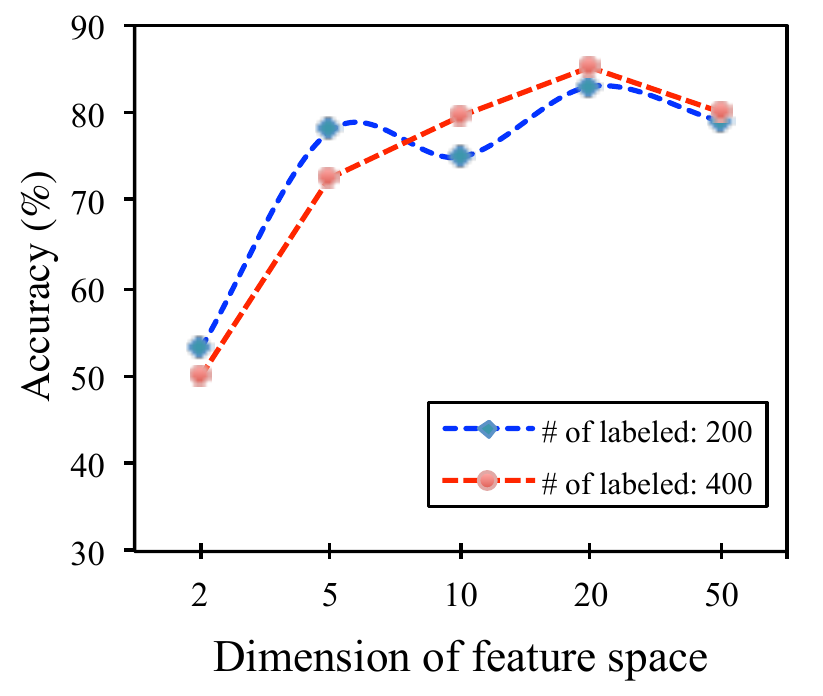}
\caption{Noisy MNIST}
  \label{fig:mnistS}
\end{subfigure}%
\begin{subfigure}{.38\textwidth}
  \centering
  \includegraphics[width=.85\linewidth]{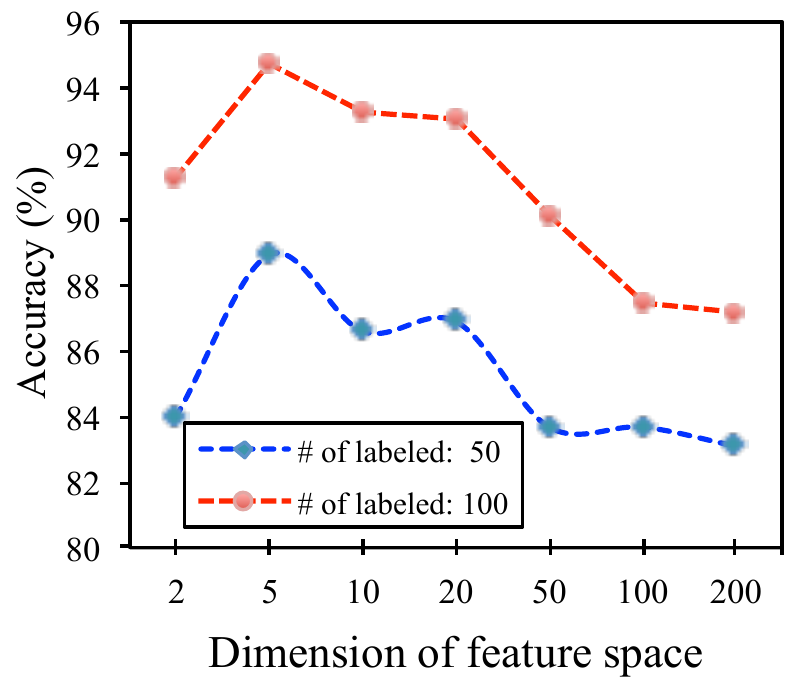}
  \caption{WebKB}
  \label{fig:wk}	
\end{subfigure}
\begin{subfigure}{.38\textwidth}
  \centering
  \includegraphics[width=.85\linewidth]{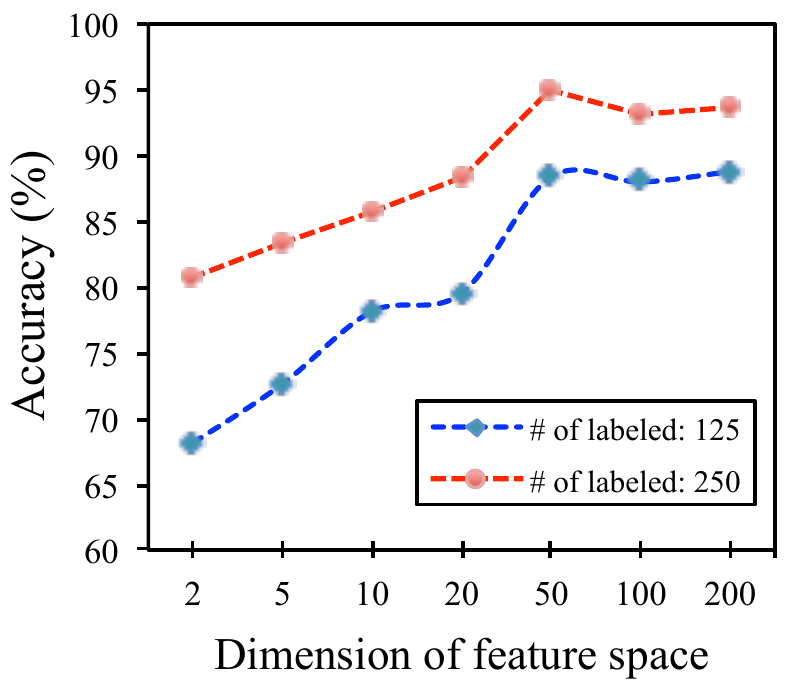}
  \caption{FOX}
  \label{fig:fox}
\end{subfigure}
\begin{subfigure}{.38\textwidth}
  \centering
  \includegraphics[width=.85\linewidth]{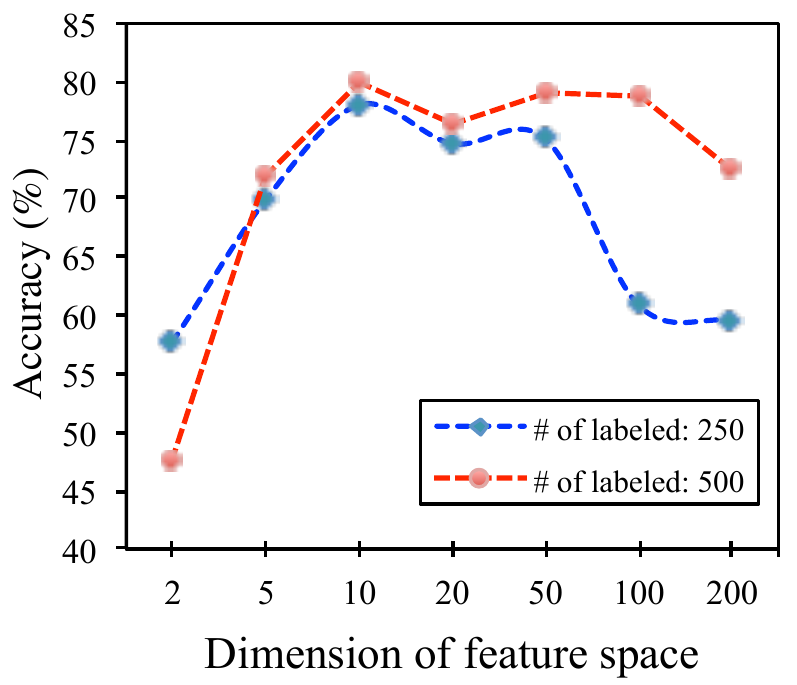}
  \caption{CNN}
  \label{fig:cnn}
\end{subfigure}
\end{center}
\caption{Accuracy of MDNN trained with two different numbers of labeled samples and various feature space size on (a) Noisy MNIST, (b) WebKW, (c) FOX and (d) CNN.}
\label{fig:acc_with_size}
\end{figure*}

\subsection{Performance Evaluation}
\label{subsec:pe}
We evaluate the effectiveness of the new representations learned by MDNN on cross-view classification tasks. All the results are reported for the primary view which is the only available view during the test. The classification accuracy of different baselines on datasets noisy MNIST, WebKB, FOX, and CNN are reported in Tables \ref{tab:mainres1} and \ref{tab:mainres2}. Results are reported for different numbers of labeled samples to show the effectiveness of the proposed algorithm in semi-supervised settings. The column labeled as `All' indicates the case where the label information for all samples is available. The best performance for each case is shown in bold. As it can be observed, MDNN outperforms all the other baselines in most cases.

The differences of MDNN's accuracies are more significant comparing to others in cases with fewer labeled samples. It shows the effectiveness of the proposed algorithm in exploiting labeled information which helps the model in semi-supervised settings. It should be considered that none of the current approaches can exploit both the labeled and unlabeled data together.

The experiments also demonstrate that MDNN can also show superior results even for supervised settings where all data are labeled. It shows that the idea of combining inter-view correlation and intra-view discrimination can be effective even when label information is available for all samples.

MDNN demonstrates better accuracy compared to deep CCA because it considers both label information and cross-view correlation when finding the projections; while deep CCA ignores the available label information. The proposed MDNN attempts to produce more discriminative feature sets by leveraging label information into the mapping learning process. Simultaneous optimization of inter-class separation, intra-class variation, and cross-view correlation make the new representations more discriminative; therefore, prediction is easier. 

Kernel CCA shows better results than MDNN in some cases of noisy MNIST. It can be due to the simplicity of noisy MNIST dataset. As it can be seen, just a few labeled samples are enough to get good results on this dataset.

\subsection{Model Analysis}
\label{subsec:ma}
We investigate and explore the influence of the main parameter of MDNN, the size of new representation, on the classification task. In Fig.~\ref{fig:acc_with_size}, the accuracies of MDNN on all datasets are plotted for various sizes of space. As it can be seen, good results can get achieved with a small size of representation, and there is no need to learn a high dimensional space. A simple classification algorithm such as linear SVM can classify the samples in the new space efficiently. It shows the representation learning power of MDNN. Representation learning can make it feasible to work on high dimensional data for the algorithms which are not able to handle high dimensional data efficiently.

Additionally, it can be seen that having unnecessary large sizes for the output dimension can affect the performance. For most datasets, hidden output size close to the number of classes can be a good choice. Unnecessary large embedding size may reduce the performance. It can be the result of producing noisy information in higher dimensional space.

\subsection{Subspace Analysis}
\label{sebsec:sa}
\begin{figure}[h!]
    \centering
    \includegraphics[width=0.45\textwidth]{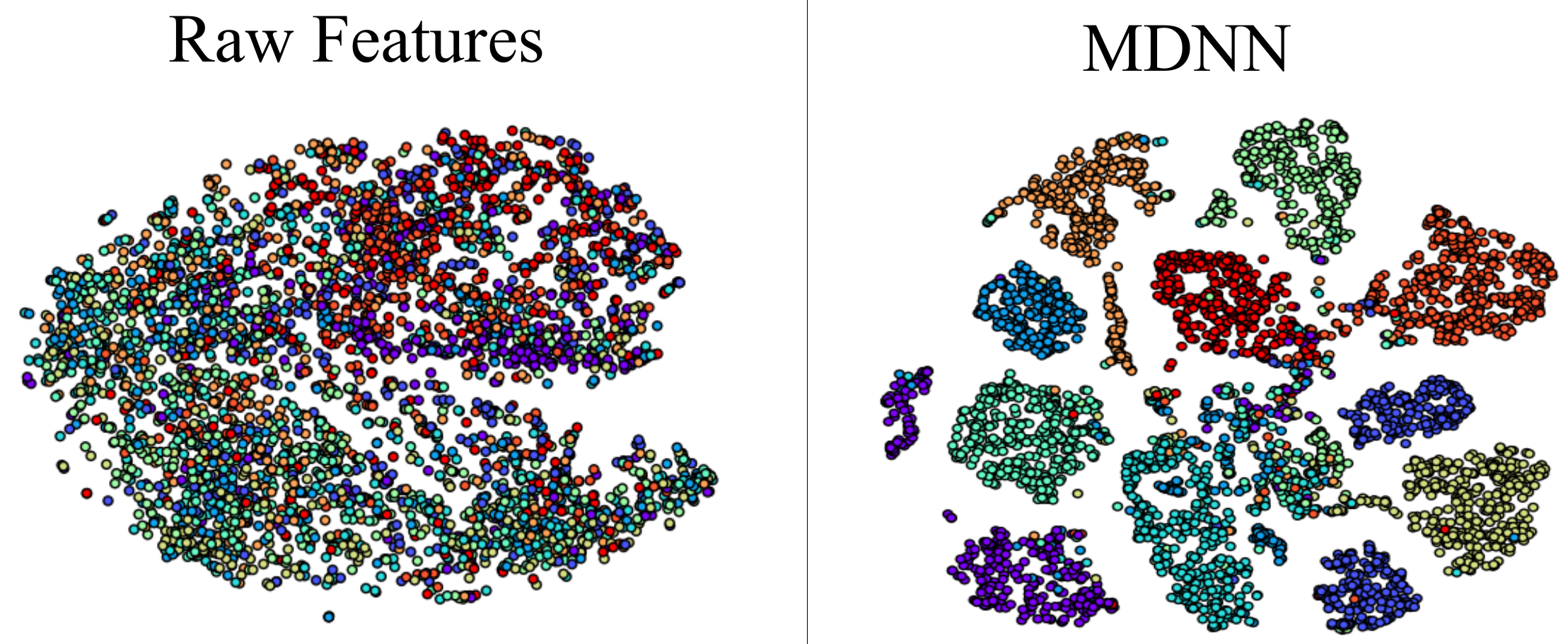}
\caption{Randomly selected instances from Noisy MNIST dataset are visualized in a 2-dimensional space using \textit{t}-SNE. They are mapped to the new space learned by MDNN. Color of the instances shows their class.}
\label{fig:tsne} 
\end{figure}

In this section, the new subspace learned by MDNN is investigated and compared with the original feature space. 4000 instances of the training samples with new representation are selected randomly and visualized in 2-dimensional space in Fig.~\ref{fig:tsne}. They are visualized using a dimensionality reduction algorithm called t-distributed stochastic neighbor embedding (\textit{t}-SNE) algorithm \cite{tsne}. It is an unsupervised representation learning that is mostly used for visualizing features in a low-dimensional space. It learns mappings from the given feature space to a new space in which similarity of samples is preserved as much as possible. In other words, samples which are close or similar in the source feature space are likely to be close to each other in the new space. It is evident that MDNN produces a more discriminative space comparing to original feature space. It learns better representation that is owed to exploiting the label information. 

\section{Conclusion}
\label{sec:conc}
We have proposed a semi-supervised deep neural network model, called MDNN, to learn discriminative representations for multi-view problems when labels for some instances are not available. To achieve this, the proposed model maximizes between-class separation and minimizes within-class variation to make the new space discriminative. It also maximizes the correlation between all views to exploit the inter-view information and also exploit the information in unlabeled data. Our model is capable of exploiting the information in both the labeled and unlabeled data in a unified learning process.

To the best of our knowledge, the proposed MDNN is the first deep network model that learns a common subspace with such properties for semi-supervised multi-view problems. The experimental results demonstrated the effectiveness of MDNN in learning discriminative feature spaces and also benefiting from the information exists in the unlabeled data.
\bibliographystyle{IEEEtran}
\bibliography{refs.bib}

\end{document}